\documentclass[11pt,a4paper]{article}
\usepackage[hyperref]{emnlp2020}
\usepackage{times}
\usepackage{latexsym}
\usepackage{graphicx}
\usepackage{subfigure}
\usepackage{amsmath}

\usepackage{microtype}

\aclfinalcopy % Uncomment this line for the final submission
\title{AutoRC: Improving BERT Based Relation Classification Models via Architecture Search}

\author{
Wei Zhu$^1$ \quad Xipeng Qiu$^2$ \quad Yuan Ni$^3$ \quad Guotong Xie$^{3,4,5}$ \\
$^1$ East China Normal University \\
$^2$ School of Computer Science, Fudan University  \\
$^3$ Pingan Health Tech \\
$^4$ Ping An Health Cloud Company Limited \\
$^5$ Ping An International Smart City Technology Co., Ltd \\
52205901018@stu.ecnu.edu.cn, xlwang@cs.ecnu.edu.cn, \\
xpqiu@fudan.edu.cn, \{niyuan442, xieguotong\}@pingan.com.cn \\
}

\date{}

\begin{document}
\maketitle
\begin{abstract}
Although BERT based relation classification (RC) models have achieved significant improvements over the traditional deep learning models, it seems that no consensus can be reached on what is the optimal architecture. Firstly, there are multiple alternatives for entity span identification. Second, there are a collection of pooling operations to aggregate the representations of entities and contexts into fixed length vectors. Third, it is difficult to manually decide which feature vectors, including their interactions, are beneficial for classifying the relation types. In this work, we design a comprehensive search space for BERT based RC models and employ neural architecture search (NAS) method to automatically discover the design choices mentioned above. Experiments on seven benchmark RC tasks show that our method is efficient and effective in finding better architectures than the baseline BERT based RC model. Ablation study demonstrates the necessity of our search space design and the effectiveness of our search method.\footnote{The source code will be made public available.}

\end{abstract}

\section{Introduction}

The task of relation classification (RC) is to predict semantic relations between pairs of entities inside a context. Given a sequence of text (usually a sentence) $x$ and a pair of entities,  $e_1$ and $e_2$, the objective is to identify
the relation between the two entities \cite{semeval2010task8}. For example, the sentence “The [kitchen]$_{e_1}$ is the last renovated part of the [house]$_{e_2}$.” shows the Component-Whole relation between the entities, “kitchen” and “house”. It is an important NLP task since it serves as an intermediate step in variety of NLP applications. There are many works applied deep neural networks (DNN) to relation classification \cite{Socher2012Semantic,zeng2014rccnn,shen-huang-2016-attention}. With the rise of pre-trained language models (PLMs) \cite{devlin2018BERT}, a series of literature have incorporated PLMs such as BERT in RC tasks \cite{baldini-soares-etal-2019-matching,enrich2019alibaba,spert,peng2019transfer}, and shows significant improvements over the traditional DNN models.  

Despite great success, there is yet no consensus reached on how to represent the entity pair and their contextual sentence for a BERT based RC model. First, in terms of entity identification in the input tokens, \citet{baldini-soares-etal-2019-matching} finds that adding entity markers around the two entities brings performance improvements, while \citet{peng2019transfer} chooses to replace entity mentions with special tokens. Second, aggregation of entity representations and contexts. \citet{baldini-soares-etal-2019-matching} is in favor of using the representations of the starting tokens of the entities as representations for the entities, however, \citet{enrich2019alibaba} prefers average pooling operations. Third, choosing which features should be considered is difficult and troublesome for manually model tuning. \citet{baldini-soares-etal-2019-matching} uses the features of the two entities, while \citet{enrich2019alibaba} incorporate the features for the [CLS] token. \citet{spert} finds that the context between the two entities is a useful signal for RC. In addition, previous literature does not consider the interactions between the feature vectors. 

In this work, we experiment on making the design choices in the BERT based RC model automatically, so that one can obtain an architecture that better suits the task at hand. Firstly, a comprehensive search space for the design choices that should be considered in a BERT based RC model is established. Second, to navigate on our search space, we employ reinforcement learning (RL) strategy. That is, a decision maker generates new RC architectures, receives rewards, and updates its policy via policy gradient method. To speed up the search procedure, shared parameters, including the parameters for BERT, pooling operations, the classification layer, are used to initialize the generated child models. To avoid over-fitting, we maintain multiple copies of BERT encoder during search. Experiments on seven benchmark RC tasks show that our method can outperform the standard BERT based RC model significantly. Transfer of the learned architecture across different tasks is investigated, which suggests tasks of different domains indeed requires task specific models. Ablation study of the search space demonstrates the validity of the search space design. In addition, we show that maintain multiple copies of BERT encoder during search helps to improve the search results.

The contributions of the paper can be summarized as: 
\begin{itemize}
	\item We develop a comprehensive search space and improve the BERT based RC models.  
	
	\item As far as we know, we are the first to introduce NAS for BERT based models. Our search method for improving search results are universally applicable.
\end{itemize}

\section{Related Work}

Our work is closely related to the literature on neural architecture search (NAS). The field of NAS has attracted a lot of attentions in the recent years. The goal is to find automatic mechanisms for generating new neural architectures to replace conventional handcrafted ones, or automatically deciding optimal design choices instead of manually tuning \cite{bergstra2011algorithms}. Recently, it has been widely applied to computer vision tasks, such as image classification \cite{cai2018proxylessnas}, semantic segmentation~\cite{Liu_2019_CVPR}, object detection~\cite{Ghiasi_2019_CVPR}, super-resolution~\cite{ahn2018fast}, etc. However, NAS is less well studied in the field of natural language processing (NLP), especially in information extraction (IE). Recent works~\cite{Zoph17a,enas,Liu18} search new recurrent cells for the language modeling (LM) tasks. The evolved transformer~\cite{so2019evolved} employs an evolution-based search algorithm to generate better transformer architectures that consistently outperform the vanilla transformer on 4 benchmark machine translation tasks. In this work, we design a method that incorporate NAS to improve BERT based relation extraction models.

Our work is closely related to relation extraction literature, especially the recent ones that take advantages of the pre-trained language models (PLMs). In terms of entity span identification, \citet{baldini-soares-etal-2019-matching} argues that adding entity markers to the input tokens works best, while \citet{peng2019transfer} shows that some RC tasks are in favor of replace entity mentions with special tokens. For feature selection, \citet{baldini-soares-etal-2019-matching} shows that aggregating the entity representations via start pooling works best across a panel of RC tasks. Meanwhile, \citet{enrich2019alibaba} chooses average pooling for entity features. In addition, it argues that incorporating the representation of the [CLS] token is beneficial. \citet{spert} shows that the context between two entities serves as a strong signal on some RC task. In this work, we provide a more comprehensive overview of the design choices in BERT based RC models, and provide a solution for efficient and task-specific architecture discovery, thus alleviating NLP practitioner in the field of RE from manually or simple heuristic model tuning.

\section{Search space for RC model}
\label{sec:search space}

\begin{figure}[tb!]
\centering
\includegraphics[width=0.45\textwidth]{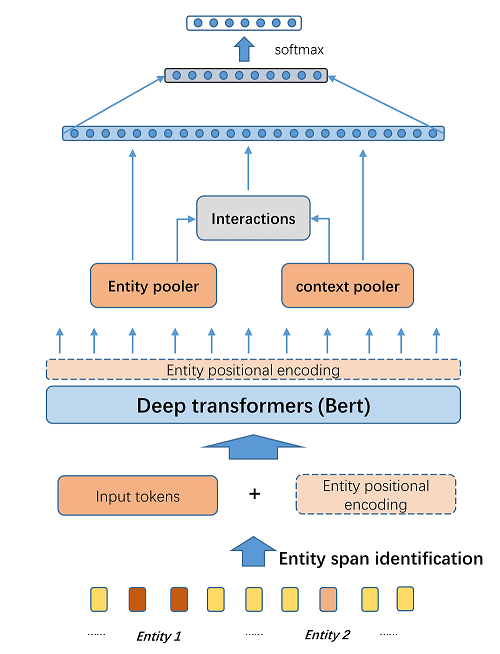}
\caption{\label{fig:autore_meta_architecture}General architecture for a RC model.}
\end{figure}

An overall architecture design for a RC model is shown in Figure~\ref{fig:autore_meta_architecture}. Following its bottom-up workflow, we will formally introduce the task  at hand and the design choices indicated in each step. Finally, we will define the search space for AutoRC.

\subsection{Formal definition of task}

In this paper, we focus on learning mappings from relation statements to relation representations. Formally, let $x = [x_0, ..., x_n]$ be a sequence of tokens, and entity 1 ($e_1$) and entity 2 ($e_2$) to be the entity mentions, which is depicted at the bottom of Figure ~\ref{fig:autore_meta_architecture}. The position of $e_i$ in $x$ is denoted by the start and end position, $s_i = (e_i^s, e_i^e)$. A relation statement is a triple $r = (x, e_1, e_2)$. Our goal is to learn a function $h_r = f_{\theta}(r)$ that maps the relation statement to a fixed-length vector $h_r \in R^d$ that represents the relation expressed in $r$. 

Note that the two entities divide the sentence into five parts, $e_1$ and $e_2$ as entity mentions, and three contextual pieces, denoted as $c_0$, $c_1$ and $c_2$.

\subsection{Pre-trained Model BERT}

In this work, we employ BERT as the encoder for the input sentences. The pre-trained BERT model \cite{devlin2018BERT} is a multi-layer bidirectional Transformer \cite{vaswani2017attention} encoder. The design of input representation of BERT is to be able to represent both a single text sentence and a pair of text sentences in one token sequence. The input representation of each token is constructed by the summation of the corresponding token, segment and position embeddings. "[CLS]" is appended to the beginning of each sequence as the first token of the sequence. The final hidden state from the Transformer output corresponding to the first token is used as the sentence representation for classification tasks. In case there are two sentences in a task, "[SEP]" is used to separate the two sentences. Thus, working with BERT encoder, sentence $x$ becomes $[[CLS], x_0, ..., x_n, [SEP]]$.

\subsection{Entity span identification}

\begin{figure}[tb!]
\centering
\vspace{-10pt}
\subfigure[standard input]{%
\includegraphics[width=0.45\textwidth]{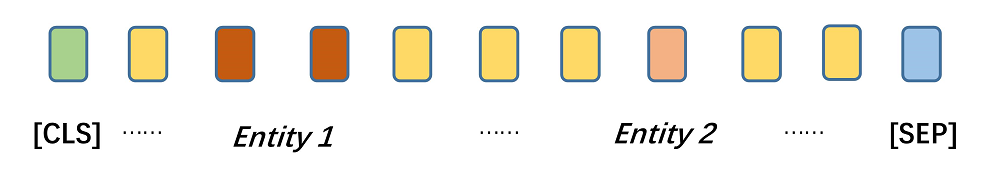}
\label{subfig:input_standard}}
\vspace{-10pt}
\subfigure[entity\_markers]{%
\includegraphics[width=0.45\textwidth]{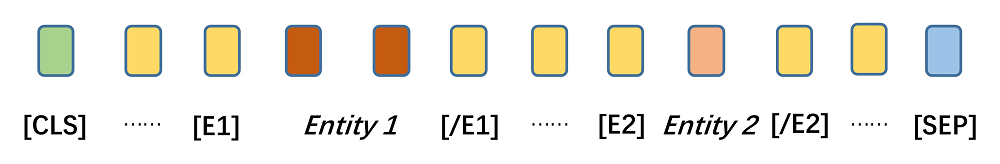}
\label{subfig:input_entity_markers}}
\vspace{-10pt}
\subfigure[entity\_tokens]{%
\includegraphics[width=0.45\textwidth]{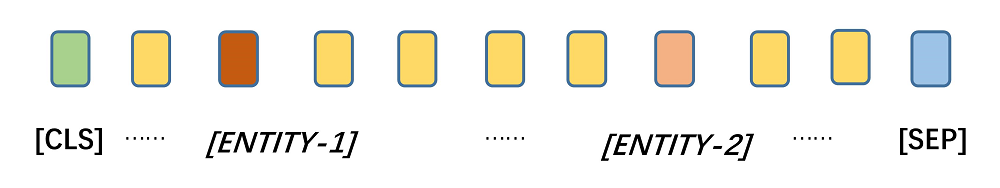}
\label{subfig:input_entity_tokens}}

\caption{How to make changes to the input sequence for entity span identification.}
\label{fig:input_formats}
\end{figure}

The BERT encoder may need to distinguish the entity mentions from the context sentence, to properly modeling the semantic representations of a relation statement, to determined the relation type. We present three different options for getting information about the entity spans $s_1$ and $s_2$ into our BERT encoder, which are depicted in Figure \ref{fig:input_formats}.

\textbf{standard}, that is, not to make any change any of input sentence (Figure \ref{subfig:input_standard}), so that BERT has no way of knowing which two entity mentions are in question.

\textbf{entity\_markers}. In this approach, we add special tokens at the start and end of the entities to info BERT where the two entities are in the sentence, as depicted by Figure \ref{subfig:input_entity_markers}. Formally, we insert $[E1]$, $[/E1]$, $[E2]$, $[/E2]$ into $x$, so that $x$ becomes
\begin{equation}
\begin{aligned}
\widetilde{x} = & [[CLS], x_0 \, ... [E1], x_{e_1^s} ... x_{e_1^{e}}, [/E1] \\
& ...\, [E2], x_{e_2^s} ... x_{e_2^{e}}, [/E2]\, ...\, x_n, [SEP]]
\end{aligned}
\end{equation}

\textbf{entity\_tokens}. As depicted by Figure \ref{subfig:input_entity_tokens}, this approach replaces the entity mentions in the sentence with special tokens. Formally, with this approach, $x$ becomes 
\begin{equation}
\begin{aligned}
\widetilde{x} = & [[CLS], x_0 \, ... \, x_{e_1^s-1}, [ENTITY-1], x_{e_1^s+1} \\
& ...\, x_{e_2^s-1},[ENTITY-2], x_{e_2^{e}+1}\, ...\, x_n, [SEP]]
\end{aligned}
\end{equation}

\subsection{Entity positional encoding}

\begin{figure}[tb!]
\centering
\includegraphics[width=0.50\textwidth]{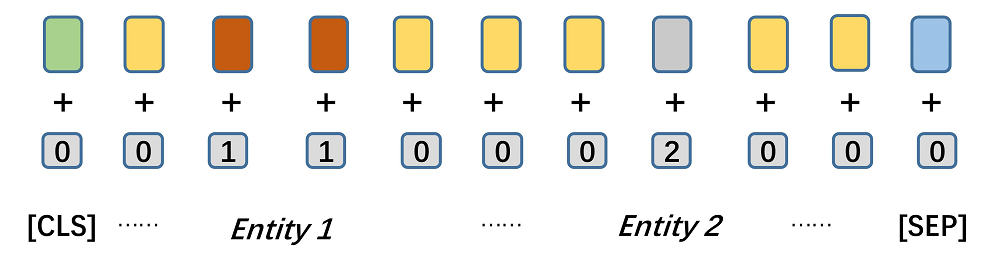}
\caption{\label{fig:entity_positional_encoding} An example of the entity positional encoding.}
\end{figure}

To make up for the standard input's lack of entity identification, or to further address the position of entities, one can add a special entity positional encoding accompany input sequence $x$. As is shown in Figure~\ref{fig:entity_positional_encoding}, the entity positional encoding will be 1 at the positions of $e_1$ and 2 at the positions of $e_2$, while the other tokens in the contextual sentence will be given 0. This entity positional encoding corresponds to an entity positional embedding module in the RC model, and it is randomly initialized and fine-tuned during BERT fine-tuning. 

Now there are two design choices. First is whether to use entity positional encoding at all. Second, as is shown in Figure~\ref{fig:autore_meta_architecture} if using entity positional encoding, do we add this extra embedding to the embedding layer of the BERT (denoted as add\_to\_embedding), or do we concatenate this embedding to the output of BERT encoder (denoted as concat\_to\_output)?

\subsection{Pooling layer}

% \begin{figure}[tb!]
% \centering
% \subfigure[entity pooler]{%
% \includegraphics[width=0.50\textwidth]{entity_pooler.PNG}
% \label{subfig:entity_pooler}}
% \subfigure[context pooler]{%
% \includegraphics[width=0.50\textwidth]{context_pooler.PNG}
% \label{subfig:context_pooler}}
% \vspace{-10pt}
% \caption{poolers for entities and contexts.}
% \label{fig:input_formats}
% \end{figure}

The output of the BERT model are the representations of the whole sentence, and its [CLS] token. Thus, how to aggregate the entities and contexts into fixed length feature vectors, i.e., what kind of poolers are used becomes the core part of the RC model architecture. In this work, we investigate how to select different pooling operations for different parts of the sentence. The pooling operations we consider includes: average pooling (avg\_pool), max pooling (denoted as max\_pool), self-attention pooling (denoted as self\_attn\_pool), dynamic routing pooling (dr\_pool) (\cite{gong2018information}), and start pooling (start\_pool), which is to use the reprsentation of the starting token as in \citet{baldini-soares-etal-2019-matching}. \footnote{Note that the representation of $[CLS]$ is used when context $c_0$ select start\_pool as the pooling operation. }

\subsection{Output features}

% \begin{figure}[tb!]
% \centering
% \subfigure[interaction between entities]{%
% \includegraphics[width=0.50\textwidth]{interaction_% entities.PNG}
% \label{subfig:interaction_between_entities}}
% \subfigure[interaction between entity and % % % context]{%
% \includegraphics[width=0.50\textwidth]{interaction_entity_and_context.PNG}
% \label{subfig:interaction_entity_and_context}}
% \vspace{-10pt}
% \caption{interactions among entities and contexts.}
% \label{fig:input_formats}
% \end{figure}

To select appropriate features for classifying relation types, there are many design choices. First, whether the two entity vectors should be used as features. Second, whether each contextual piece ($c_0$, $c_1$, $c_2$) should be added as features \cite{spert,enrich2019alibaba}. 

We notice that the literature does not consider the interactions of the features from different parts of the sentence, which proves to be useful in other tasks such as natural language inference (NLI) \cite{chen2016enhanced}. Here, we consider the interaction between the two entities, and their interactions with contextual pieces. The interaction can be dot product (denoted as dot) or absolute difference (denoted as minus) between two feature vectors.

\subsection{Search space}

Now we are ready to define the search space formally. The search space is as follows:

\begin{itemize}
    \item how to distinguish entities from the contexts = {entity\_markers, entity\_tokens, standard};
    \item whether and how to use entity positional embedding = {null, add\_to\_embedding, concat\_to\_output}, where null means no interaction;
    \item pooling operations for entity or contextual piece = {avg\_pool, max\_pool, self\_attn\_pool, dr\_pool, start\_pool};
    \item whether to use the representation of entity $e_i$ = {True, False}, where $i=1,2$;
    \item whether to use the representation of context $c_i$ = {True, False}, where $i=0,1,2$;
    \item Interaction between the two entities = {dot, minus, null}, where null means no interaction;
    \item Interaction between entity and contextual piece $c_i$ = {dot, minus, null}, where null means no interaction, and $i=0,1,2$.
    
\end{itemize}

Our search space contains 1.64e+8 combinations of design choices, which makes manually fine-tuning or random search impractical. 

\section{Search method}

In this section, we elaborate on how to search on the search space defined in the previous section.

\subsection{LSTM decision maker}

In AutoRC, an decision maker is used to construct the architecture for a BERT based RC model. Denote T as the number of design choices required, and each design choice is denoted as $e_t$ ($t=0, 1, ..., L - 1$), and $e_t$ has $m_t$ possible values ${y_t^{0}, ..., y_t^{m_t-1}}$. Our aim is to decide the values for these design choices using this decision maker. The decision maker uses a single-layer LSTM model (Hochreiter and Schmidhuber, 1997) as a
backbone and sequentially predicts the value for each design choice.

More formally, the input to the decision maker is a special token “[Start]”, and the model proceeds to predict the value for each design choice. At each time step, the input to the LSTM
model is the representation $h_t$ for the node that has just been predicted. Then it is combined with
the representation previously constructed representation $g_t$ to obtain $g_{t+1}$ representing the current
time-step using LSTMs, and $h_t$ is fed into a linear layer $f_t$ and softmax function to decide the value for design choice $e_t$. That is, the probability of design choice $e_t$ being equal to $y_t^{l}$ is 
\begin{equation}
    P(e_t = y_t^{l}) = \frac{\exp(f_t(h_t^l))}{\sum_{l^{'}} \exp( f_t(h_t^{l^{'}})) }
\end{equation}

\subsection{Search and evaluation}

The whole procedure for our method can be divided into the search phase and evaluation phase. The search phase updates the shared parameters and the parameters for the decision maker in an interleaving manner, while the evaluation phase obtains multiple top-ranked models from the decision maker and train them till convergence on the task dataset for proper evaluations of the learned architectures.   

\textbf{Search phase.} The search phase aim at training the decision maker so that it can generate better architectures than the baseline. Directly training the decision maker is impractical since we have no access to the ground-truth design choices. We use REINFORCE \cite{williams1992simple}, which is an instance of a broader class of policy gradient methods for optimization. The main idea is to use reinforcement learning to discover the best combination of design choices for the RC task. 

Denote each action as $a$, which is the architecture predicted by the decision maker. Let $\Theta$ denote parameters of the decision maker and $\Phi$ denote the shared parameters of the main network. Then the policy of the decision maker is denoted as $\pi(a, \Theta)$. Under the framework of reinforcement learning, we ask the decision maker to maximize its reward $R(\Theta)$, which can be the metric specified by the RC task at hand, e.g., micro/macro F1. 

Now we describe the interleaving optimization procedure. First, an architecture is sampled by the decision maker, and its network parameters are initialized with $\Phi$. It is trained for $n_c$ epoch (which is usually a small number like 0.5 or 1), during which $\Phi$ is updated. Then, $N$ architectures are generated by the decision maker and their rewards are obtained when each architecture is initialized with $\Phi$. With the reward signal, $\Theta$ is updated using policy gradients: 
\begin{equation}
\nabla_\Theta \hat{J}(\Theta) = \frac{1}{N}\sum_{i=1}^{N}\nabla_\Theta \log \pi(a_i, \Theta) (R(\Theta) - b),
\end{equation}
where $b$ denotes a moving average of the past rewards and it is used to reduce the variance of gradient approximation. In this work, we find $N=1$ already works quite well. Repeating this interleaving optimization procedure for $n_d$ times till the decision maker is well trained, then we generate $N_f$ candidate architectures, evaluate them using the shared parameters, and then select the top-ranked $N_e$ models for architecture evaluation. 

\textbf{Evaluation phase.} In this phase, the top-ranked models are trained with the whole train set, and validated on the dev set to select the best checkpoint for prediction on the test set. Note that the shared parameters $\Phi$ are discarded in this phase, and the learned architecture is trained from scratch. To fully evaluate each architecture, we run a grid search for the optimal hyper-parameters including learning rate, batch size and warm-up steps. After the optimal combination of hyper-parameters is selected, the model is run several times to ensure replication.

\section{Experiments}

\subsection{Datasets}

\begin{table}
\centering
		
	\resizebox{0.48\textwidth}{!}{
		\begin{tabular}{cccccccc}
			\hline
			\bf Dataset & \bf \# labels &\bf Train &\bf Dev & \bf Test & \bf  sent length & \bf Metrics  \\ 
			\hline
			semeval2010 & 19 & 6508 & 1494 & 2718 & 19.09 & micro F1  \\
			kbp37 & 37 & 15917 & 1724 & 3405 & 31.09 & micro F1  \\
			wiki80 & 80 & 40320 & 10080 & 5600 & 24.93 & micro F1  \\
			deft2020 & 6 & 16727 & 963 & 1139 & 72.11 & macro F1  \\
			i2b2 & 8 & 2496 & 624 & 6293 & 24.33 & micro F1  \\
			ddi & 5 & 18779 & 7244 & 5761 & 45.03 & micro F1  \\
			chemprot & 6 & 19460 & 11820 & 16943 & 49.69 & micro F1  \\
			\hline
		\end{tabular}}
		
		\caption{\label{tab:dataset-statistics}Overview of datasets in experiments.}
\end{table}

We run experiments on 7 different datasets, which are from various domains and are different in the respects of dataset sizes, sentence length, entity mention length, etc, to demonstrate that our method is robust for various RC tasks. 

\textbf{SemEval-2010 Task 8} \cite{semeval2010task8} (denoted as semeval10) This dataset does not establish a default split for development, so for this work we adopt the same train/dev split with that provided by OpenNRE \cite{han-etal-2019-opennre}.

\textbf{Wiki80 } (denoted as wiki80) This dataset \cite{han-etal-2019-opennre} is
derived from FewRel \cite{han-etal-2018-fewrel}, a large
scale few-shot dataset. Since Wiki80 only has a train/val split, we randomly split the train set into a train set and val set (with 8:2 ratio), and treat the original validation set as the test set.

\textbf{KBP-37} \cite{re_by_rnn} (denoted as kbp37).  This dataset is a revision of MIML-RE annotation dataset, provided by Gabor Angeli et al. (2014). They use both the 2010 and 2013 KBP official document collections, as well as a July 2013 dump of Wikipedia as the text corpus for annotation.

\textbf{DEFT-2020 Subtask 3} (denoted as deft2020) This dataset also serves as the task 6 of SemEval 2020 shared tasks. This RC task have to overcome longer contexts, longer entity mentions, and more imbalanced relation types. \cite{spala-etal-2019-deft}

\textbf{i2b2 2010 } (denoted as i2b2) shared task collection consists of 170 medical documents for training and 256 documents for testing, which is the subset of the original dataset \cite{Uzuner20112010}.

\textbf{ChemProt} (denoted as chemprot) consists of 1,820 PubMed abstracts with chemical-protein interactions annotated by domain experts and was used in the BioCreative VI text mining chemical-protein interactions shared task \cite{krallinger2017overview} \footnote{https://biocreative.bioinformatics.udel.edu/news/corpora/}. 

\textbf{DDI} extraction 2013 corpus (denoted as ddi) is a collection of
792 texts selected from the DrugBank database
and other 233 Medline abstracts \cite{herrero2013ddi}.\footnote{http://labda.inf.uc3m.es/ddicorpus}

\subsection{Search protocol}
\label{subsec:arch_search_protocol}

During search phase, the interleaving optimization process is run 100 times. For each search epoch, a proportion $r=0.5$ of the train data is passed to a child model. Throughout this work, we use the base uncased version of BERT \cite{devlin2018BERT} as the sentence encoder, and its parameters are fine-tuned to better adjust to downstream tasks. During search, to avoid over-fitting and unreliable approximations of a sampled model, we maintain 3 copies of BERT model checkpoints, so each time we initialize a child model, a BERT checkpoint is randomly selected and its parameters can be updated. If the entity position embedding is concatenated after the BERT output, its size is set to be 12. 

During search, the learning rate for the decision maker is set at 1e-4, and the learning rate and batch size for the sampled architectures are manually tuned to obtain better search results. During search, the number of warm-up steps is set to be equal to 0.3 of a epoch.  

After the search phase, 30 model architectures are sampled from the trained controller, and they are ranked via their performance on the valid data when they are initialized using the shared parameters. Then the top-ranked 5 models are trained from scratch till convergence on the whole training data of the task to formally evaluate their performances.

To compare our methods with random search, for each task, we randomly samples 7 different models with a randomly initialized decision maker, since the GPU time for training 7 models is guaranteed to be larger than an entire search and evaluation process described above. We train them from scratch, and report the performance of the best generated model as the performance of this random search run. The results of random search will be the average of 3 such runs. 

Due to resource limitations, we assign up to 2 NVIDIA V100 GPU cards to each tasks. 

\begin{table*}
\centering
\resizebox{0.98\textwidth}{!}{
\begin{tabular}{cccccccc}
\hline
\textbf{Model} & \textbf{semeval10} & \textbf{kbp37} & \textbf{wiki80} & \textbf{deft2020} & \textbf{i2b2} & \textbf{ddi}  & \textbf{chemprot} \\
\hline
BERT-entity & 88.3$^{*}$ & 64.20 $\pm$ 0.273 &  85.35 $\pm$ 0.141 & 60.19 $\pm$ 0.816 &  81.94 $\pm$  0.691  & 75.66 $\pm$  0.712 &  66.86 $\pm$ 0.393 \\
\hline
random search  & 87.61 $\pm$ 0.316 & 63.90 $\pm$ 0.516 &  83.46 $\pm$ 0.378 & 58.19 $\pm$ 1.968 &  81.33 $\pm$  1.364  & 74.23 $\pm$  0.653 &  66.04 $\pm$ 0.873 \\
$AR_{semeval10, 0}$  & \textbf{88.55} $\pm$ 0.188 & - & - & - & - & -  & - \\
$AR_{kbp37, 1}$  & - & \textbf{64.57} $\pm$ 0.175 & 85.53 $\pm$ 0.157 & - & 81.82 $\pm$ 0.764 & 75.58 $\pm$ 0.704 & -  \\
$AR_{wiki80, 2}$  & - & - & \textbf{85.66} $\pm$ 0.116 & - & -  & - & -  \\
$AR_{deft2020, 1}$  & - & - & - & \textbf{63.58} $\pm$ 0.529 & - & - & -\\
$AR_{i2b2, 0}$   & - & - & - & - & \textbf{83.37} $\pm$ 0.456 & 75.89 $\pm$ 0.638 & -\\
$AR_{ddi, 1}$   & - & - & - & - & 82.86 $\pm$ 0.496 & \textbf{76.14} $\pm$ 0.421 & - \\
$AR_{chemprot, 0}$  & - & -  & - & - & - & - & \textbf{67.46} $\pm$ 0.271 \\
\hline
\end{tabular}}
\caption{\label{tab:main_results}
Test results for seven relation classification tasks. Results on rows where the model name is marked with a * symbol are reported as published, all other numbers have been computed by us. The performance metric is micro F1 for all tasks except for deft2020 which uses macro F1. Results from the baseline model are obtained with the help of OpenNRE \cite{han-etal-2019-opennre}.
}
\end{table*}

\subsection{Architecture evaluation protocol}
\label{subsec:arch_eval_protocol}
To thoroughly evaluate a learned model, we run a random search of 10 times on the following space for the optimal combination of the following key hyper-parameters:

\begin{itemize}
    \item learning rate = 1e-3, 1e-4, 5e-5, 2e-5, 1e-5;
    \item training batch size = 128, 64, 32, or 16;
    \item warm-up steps = 0.5, 0.8, 1.0 of the number of steps in an epoch;
    % \item dropout rate for the output of the BERT encoder = 0.1, 0.3, 0.5.
\end{itemize}

In addition, to make our results more reproducible, each learned model is trained under the optimal hyper-parameters for 10 times, and the mean and variance of the performance will be reported.

In this work, we select the so-called \textbf{BERT-entity} models from the OpenNRE framework \cite{han-etal-2019-opennre} as the baselines, since they have open-sourced their codes and are easy to implement. The results of the baseline models are special cases of our model, so their performances by the OpenNRE serves as a quality check for our codes. The baseline models also have to go through the above reproducibility protocols, except that we report the results given out by OpenNRE. We will not compare with traditional deep-learning based model in the pre-BERT era, since \textbf{BERT-entity} significantly outperforms them.   

The baseline model and the model sampled by an un-trained decision maker will also go through the same evaluation procedure.

\begin{table*}
\centering
\resizebox{0.70\textwidth}{!}{
\begin{tabular}{ccc}
\hline \textbf{Search space} & \textbf{deft2020} & \textbf{i2b2} \\ 
\hline
$SS$ & \textbf{63.58} $\pm$ 0.529 & \textbf{83.37} $\pm$ 0.456 \\
$SS-no\_inte$ & 63.14 $\pm$ 0.721 &  83.12 $\pm$ 0.533   \\
$SS-no\_inte-start\_pool$ & 62.13 $\pm$ 0.523 & 82.43 $\pm$ 0.495 \\
$SS-no\_inte-start\_pool-no\_contexts$ & 61.57 $\pm$ 0.893 & 82.08 $\pm$ 0.598  \\
\textbf{BERT-entity} & 60.19 $\pm$ 0.816 &  81.94 $\pm$  0.691 \\
\hline
\end{tabular}}
\caption{\label{tab:ablation_on_search_space} Results of ablation study on the search space. }
\end{table*}

\begin{table}
\centering
\begin{tabular}{cc}
\hline \textbf{\# copies of BERT encoder} & \textbf{i2b2}  \\ \hline
3 & \textbf{83.37} $\pm$ 0.456  \\
1 & \textbf{82.58} $\pm$ 0.614  \\
\hline
\end{tabular}
\caption{\label{tab:ablation_on_num_of_BERT_copies} Effects of maintaining multiple copies of BERT encoder during search.}
\end{table}

\subsection{Results on Benchmark datasets}

The results on the 7 benchmarks RC datasets are reported in Table~\ref{tab:main_results}. Random search means the models are sampled with an un-trained decision maker, as described in the previous sub-section. We denote each learned architecture as $AR_{task\_name, idx}$, where $idx$ denotes the rank of this learned model during the search phase.

For all seven tasks, our method successfully learns a significantly better RC model than the \textbf{BERT-entity} baseline. Note that although the search phase does not always rank the best evaluated model the first place, but according to the results, one can evaluate up to 3 models from the search phase to save computation consumption.

\begin{figure}[tb!]
\centering
\includegraphics[width=0.45\textwidth]{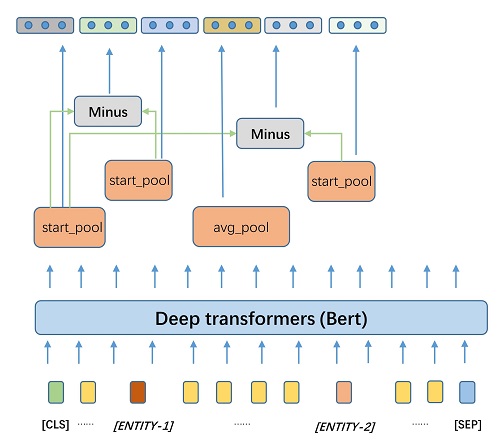}
\caption{\label{fig:deft2020_arch} Learned architecture on deft2020. }
\end{figure}

\begin{figure}[tb!]
\centering
\includegraphics[width=0.45\textwidth]{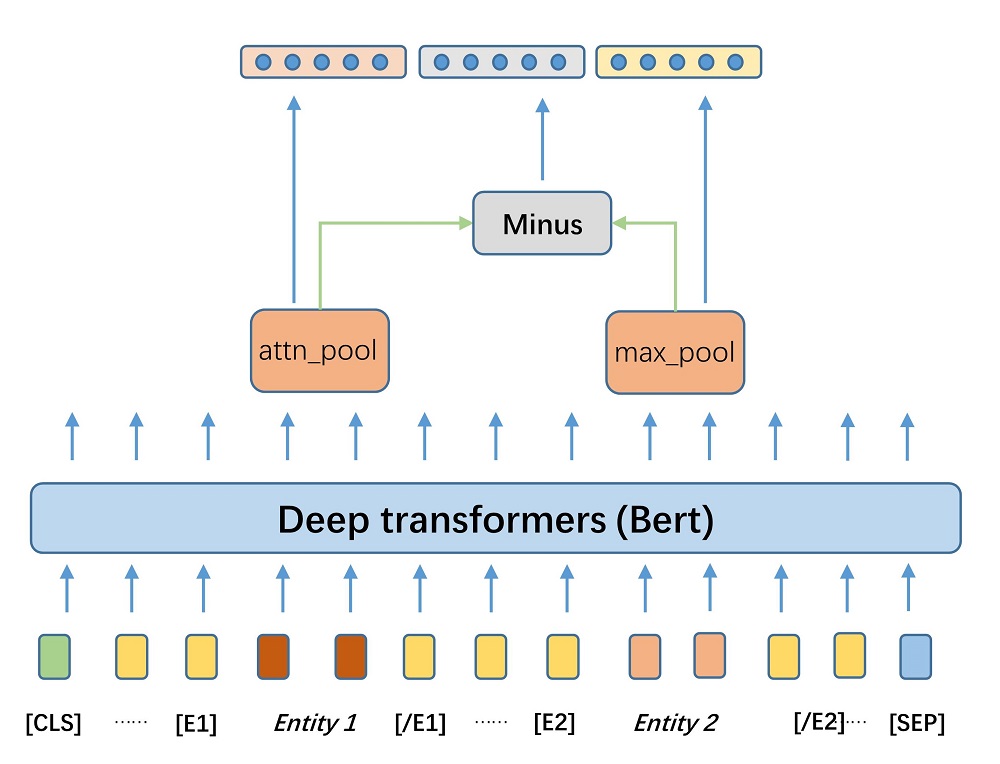}
\caption{\label{fig:i2b2_arch} Learned architecture on i2b2. }
\end{figure}

Figure \ref{fig:deft2020_arch} and \ref{fig:i2b2_arch} reports the searched architectures for the deft2020 and i2b2 tasks. We can see that learned architectures are quite different from each other, thus validating the necessity of task specificity. The two learned models are different in the following three aspects. First, $AR_{i2b2, 0}$ selects to add entity markers to indicate the position of the entities, while $AR_{deft2020, 1}$ choose to replace entity mentions with entity tokens. We hypothesis that in deft-2020, the entities are often quite long, thus replacing entity mentions with entity tokens is beneficial for the model to understand the contexts' structural patterns. Second, note that $AR_{deft2020, 1}$ uses start\_pool to aggregate context piece $c_0$, which is the representation of [CLS] token. In addition, it includes the representation of context $c_1$. Third, $AR_{deft2020, 1}$ incorporates the interaction between context $c_0$ and the two entities, while $AR_{i2b2, 0}$ include the interaction between the two entities. Differences in the learned architectures for different tasks indicate the necessity of task specific architectures, which is challenging without the help of NAS. 

One observation can be made is that the test results of the search architectures are consistently stable than the baseline, which also validates that our method are efficient in finding a task-specific model for the task at hand. 

In table~\ref{tab:main_results}, we also study how does an architecture learned on one task performs on another. Note that when evaluated on a different task, an architecture's hyper-parameters are tuned again, following the procedure described in subsection~\ref{subsec:arch_eval_protocol}. The architecture learned on kbp37, which is an open-domain dataset, $AR_{kbp37, 1}$, transfer well on wiki80. But it does not perform well on the two tasks of medical domain, i2b2 and ddi. However, the learned architectures learned on i2b2 and ddi transfer well on each other and perform comparably well. The above results demonstrate that the learned models have certain ability for task transfer, but its suitability is significantly affected by the domains of the tasks.

\subsection{Ablation study on the search space}

We have demonstrated the strong empirical results based on the proposed approach. We further want to understand the specific contributions by the different components of the search space. For this purpose, we create three smaller search space. The first one, denoted as $SS-no\_inte$, which does not allow any interactions among entity features and context features. The second one, $SS-no\_inte-start\_pool$ further reduce $SS-no\_inte$ by limiting that the pooling operation available is the start pooling operation. The third one, $SS-no\_inte-start\_pool-no\_contexts$, further forbid contextual features. If further limit the entity span identification method to be entity markers, the search space is reduced to the baseline \textbf{BERT-entity} model. The search and evaluation protocols on the reduced search space strictly follow \ref{subsec:arch_search_protocol} and  \ref{subsec:arch_eval_protocol}. 

We conduct ablation study for the search space on deft2020 and i2b2. Results are reported in Table~\ref{tab:ablation_on_search_space}. For deft2020, alternate method for span identification provides significant performance gain, and interaction among features is also important. For i2b2, the most significant performance drop occurs when the pooling operations are limited, indicating that even for powerful bi-directional context encoder like BERT, considering different pooling operations are beneficial.

\subsection{Effects of maintaining multiple copies of BERT encoder}

Now we investigate the effect of maintaining multiple BERT checkpoints during the search phase on the i2b2 task. We set the number of copies of BERT encoder as 1, and re-run the search and evaluation procedure, obtain the best architecture and report its performance on Table~\ref{tab:ablation_on_num_of_BERT_copies}. We can see that with only one copy of BERT encoder, the resulting architecture performs worse. We believe that maintaining multiple copies of BERT encoder during search can effectively avoid over-fitting, making the reward signal during search more reliable, thus resulting in better searched architectures.

\section{Conclusion}

In this work, we first construct a comprehensive search space to include many import design choices for a BERT based RC model. Then we design an efficient search method with the help of RL to navigate on this search space. To speed up the search procedure, parameter sharing is designed, including the parameters for BERT, pooling operations, the classification layer. To avoid over-fitting, we maintain multiple copies of BERT encoder during search. Experiments on seven benchmark RC tasks show that our method can outperform the standard BERT based RC model significantly. Ablation study shows our search space design is valid.

\bibliographystyle{acl_natbib}
\bibliography{emnlp2020}

\end{document}